\documentclass[10pt,twocolumn,letterpaper]{article}
\usepackage[pagenumbers]{cvpr} 
\usepackage[dvipsnames]{xcolor}
\usepackage{amsmath}
\usepackage{amssymb}
\usepackage{arydshln}
\usepackage{booktabs}
\usepackage{csvsimple}
\usepackage{multirow}
\usepackage{etoolbox}
\usepackage{graphicx}
\usepackage{makecell}
\usepackage{xspace}
\usepackage{amssymb}
\usepackage{pifont}
\definecolor{cvprblue}{rgb}{0.21,0.49,0.74}
\usepackage[pagebackref,breaklinks,colorlinks,allcolors=cvprblue]{hyperref}

\newif\ifprintcomments
\printcommentstrue

\makeatletter
\DeclareRobustCommand\onedot{\futurelet\@let@token\@onedot}
\def\@onedot{\ifx\@let@token.\else.\null\fi\xspace}
\makeatother

\def\ie{\emph{i.e}\onedot}

\newcommand{\themethod}{Twinner\xspace}
\newcommand{\xmark}{\ding{55}}%

\makeatletter
\renewcommand{\paragraph}{%
    \@startsection{paragraph}{4}%
    {\z@}{-0.5em}{-0.5em}%
    {\normalfont\normalsize\bfseries}%
}
\makeatother

\def\paperID{1117}
\def\confName{CVPR}
\def\confYear{2025}

\title{\themethod: Shining Light on Digital Twins in a Few Snaps}

\author{
\phantom{aaaaaaaaaaaa}Jesus Zarzar$^{2,3}$\and
Tom Monnier$^{3}$ \and
Roman Shapovalov$^{3}$\phantom{aaaaaaaaaaaa}\vspace{0.001cm}\and
Andrea Vedaldi$^{1,3}$ \and
David Novotny$^{3}$ \and
\\
$^1$VGG, University of Oxford \hspace{2em} 
$^2$KAUST \hspace{2em} 
$^3$Meta AI
}

\begin{document}
\twocolumn[{
\maketitle
\begin{center}
\includegraphics[width=\linewidth]{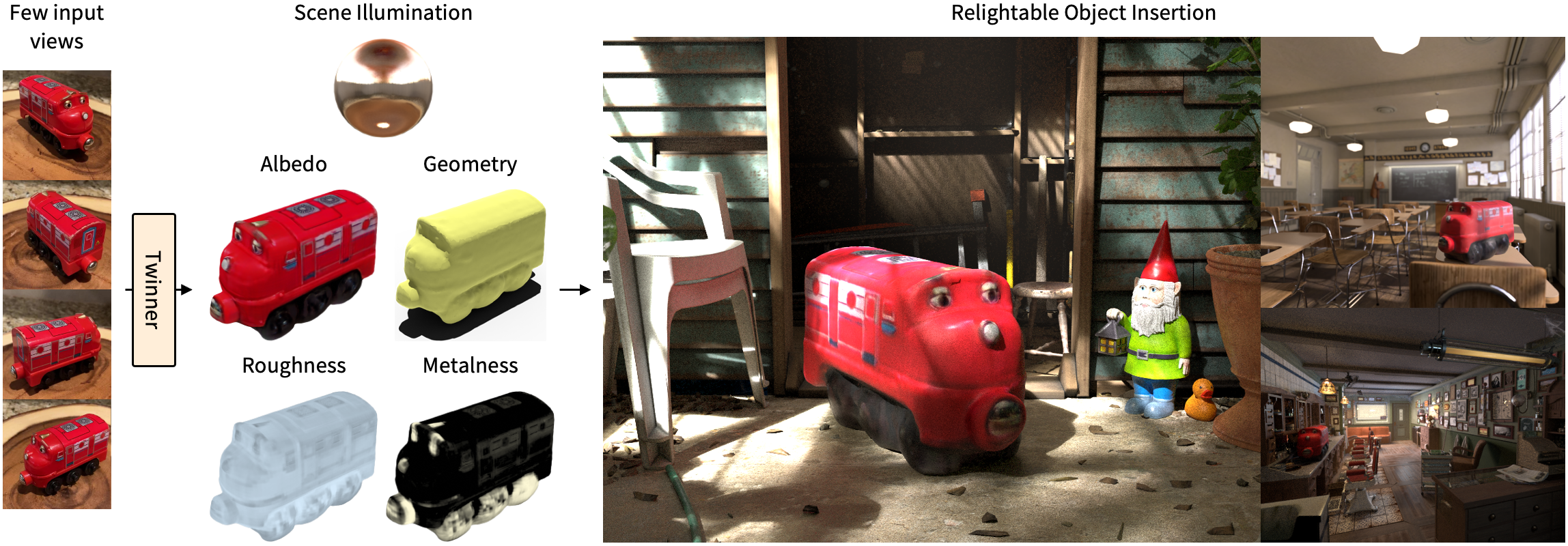}
\end{center}
\vspace{-1em}
\captionsetup{type=figure}
\captionof{figure}{We present \textbf{\themethod}, the first large reconstruction model capable of predicting a scene's illumination as well as an object's geometry and material properties from a few posed images. This enables tasks such as realistic relighting of objects in novel scenes in a few seconds by replacing costly per-scene optimizations with a single forward pass.
}%
\label{fig:teaser}
\vspace{1.75em}
}]

\begin{abstract}
We present the first large reconstruction model, \themethod, capable of recovering a scene's illumination as well as an object's geometry and material properties from only a few posed images.
\themethod is based on the Large Reconstruction Model and innovates in three key ways:
1) We introduce a memory-efficient voxel-grid transformer whose memory scales only quadratically with the size of the voxel grid. 
2) To deal with scarcity of high-quality ground-truth PBR-shaded models, we introduce a large fully-synthetic dataset of procedurally-generated PBR-textured objects lit with varied illumination. 
3) To narrow the synthetic-to-real gap, we finetune the model on real life datasets by means of a differentiable physically-based shading model, eschewing the need for ground-truth illumination or material properties which are challenging to obtain in real life. 
We demonstrate the efficacy of our model on the real life StanfordORB benchmark where, given few input views, we achieve reconstruction quality significantly superior to existing feed-forward reconstruction networks, and comparable to significantly slower per-scene optimization methods.
\end{abstract}
\section{Introduction}%
\label{sec:intro}

The creation of digital twins, \ie, the conversion of real-world objects into virtual 3D assets, is an important problem in virtual and augmented reality, gaming, conservation, education, advertisement, and general 3D content creation.
In the last few years, new techniques such as NeRF~\cite{mildenhall20nerf:} and 3D Gaussian Splatting~\cite{kerbl233d-gaussian}, along with their their feed-forward counterparts~\cite{henzler2021unsupervised,yu21pixelnerf:,tang24lgm:,xu24grm:}, have significantly improved the quality, speed, and generality of digital twinning.
However, most of theses techniques still bake illumination into the scene's appearance as radiance.
As a consequence, the resulting assets cannot be relit and look out-of-place when visualized under different lighting conditions.

To address this limitation, one must also reconstruct the field of \emph{Bidirectional Reflectance Distribution Functions} (BRDF), which describe the optical properties of the materials.
Estimating the BRDF directly is notoriously challenging due to its high dimensionality~\cite{bartell81the-theory} and is usually done by means of 3D scanners that take thousands of images controlling for viewpoint and illumination~\cite{zhang20deep,sang20single-shot,nam18practical,boss20two-shot,bi20Bdeep,bi20deep,bi20neural}.
This is often impractical, so authors have considered ``casual'' reconstruction setups, where the object is captured by a conventional camera in uncontrolled conditions.
Examples include NVDiffRec~\cite{munkberg22extracting}, NeRD~\cite{boss2021nerd} and NeRFactor~\cite{xiuming21nerfactor:}, which simultaneously estimate the illumination, in the form of an environment map, and a simplified BRDF, parameterized using \emph{Physically-Based Rendering} (PBR) with albedo, roughness and metallicity~\cite{burley12physically-based}.
These quantities are recovered by minimizing the photometric reconstruction error, comparing the output of a differentiable PBR renderer to the input images.
However, this optimization is slow and ill-posed, and requires the introduction of handcrafted priors for regularization.

In this paper, we propose a new method, \emph{\themethod}, a feed-forward network that outputs shape, PBR textures, and environment illumination.
This network is fast, performs well, requires only a small number of images as input, and learns empirically the necessary reconstruction priors.

Siddiqui et al.~\cite{siddiqui24meta} recently introduced a similar predictor, MetalLRM, as an extension of the LightplaneLRM~\cite{cao2024lightplane}.
MetalLRM requires supervision with a large dataset of 3D assets with high-quality PBR textures which, unfortunately, are not abundant in existing synthetic datasets like Objaverse~\cite{deitke23objaverse:}.
Moreover, even if a large synthetic dataset were available, the synthetic-to-real gap between the artist-created training assets and the target domain of real input images makes MetalLRM unsuitable for digital twinning.

With our new \themethod, we address the scarcity of high-quality training PBR assets and the train/test domain shift.
First, we ask whether real images in datasets like CO3Dv2~\cite{reizenstein21co3d} can be used to supervise the model.
Although these datasets lack PBR information, they are abundant and match the target distribution of real images we aim to digitally twin.
In order to use this data, we propose to minimise the photometric error similar to optimization-based models, but with the key difference that the 3D shape, PBR materials, and environment light are all \emph{predicted} by a large feed-forward transformer~\cite{hong24lrm:,siddiqui24meta}.
During training, each scene is rendered with a differentiable PBR shader that accepts the predicted environment light and the predicted PBR-textured shape.
The latter then admits indirect PBR and illumination supervision via photometric losses between the renders and the corresponding real ground-truth images.

In addition, we consider a second approach to tackle the data scarcity where, inspired by Zeroverse~\cite{xie24lrm-zero:}, we procedurally generate PBR-textured objects lit with a known set of environment maps.
While the resulting objects are not realistic, their key advantage is that they come with clean ground-truth PBR and illumination information, which can be used to supervise the corresponding predictors \emph{directly} instead of indirectly via the photometric loss.

Besides addressing data scarcity, we also improve the architecture of the reconstruction model.
We believe in the virtues of a simple and direct voxel-grid representation of the object~\cite{sun22direct}, but note that predicting a number of tokens cubic in the voxel grid resolution is intractable with a transformer.
Reducing the number of predicted tokens is the primary reason why recent models~\cite{hong24lrm:,siddiqui24meta} factor the voxel grid as a triplane.
The question, then, is how to retain the benefits of the direct voxel grid representation and still make computations feasible.
We solve this problem by introducing the novel \emph{tricolumn} representation, where each feature vector (column) in a triplane is obtained by stacking the corresponding cells in the underlying voxel grid.
In this manner, the number of tokens becomes quadratic in the voxel grid size, while each voxel is still represented explicitly.
We further augment the network to predict a representation of the environment map, capturing illumination.

We demonstrate the efficacy of \themethod model on the StanfordORB benchmark~\cite{kuang2023stanfordorb}, consisting of casual captures of real-life objects.
We demonstrate reconstruction quality superior to existing feed-forward predictors and comparable to optimization-based methods, while using only a very small number of input images of the object and operating much faster.
We also obtain state-of-the-art reconstruction of the environment maps, achieving an improvement of 28\%  in the predicted illumination's angular error over existing methods.

\begin{figure*}[htb]
\centering
\includegraphics[width=\textwidth]{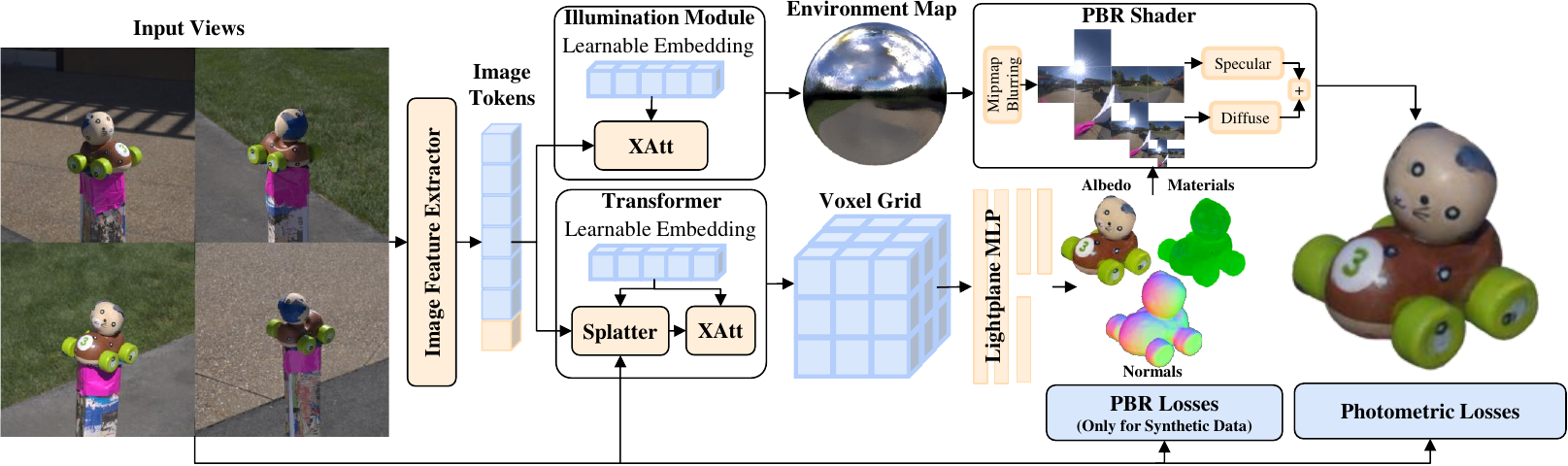}
\caption{
\textbf{Overview of \themethod.}
The input views are processed together with their foreground masks and camera poses by an image tokenizer.
The resulting tokens are processed by a Diffusion Transformer (DiT) model to predict a 3D volumetric representation of the scene, and a second DiT model to predict the scene's illumination.
The 3D representation is then rendered to obtain images of the scene's material properties, normals, and opacity from the target's point of view.
Using these rendered images together with the predicted illumination, \themethod renders an approximate physically-shaded image of the scene which then enters a photometric loss.
}%
\vspace{-1em}
\label{fig:pipeline}
\end{figure*}
\section{Related work}%
\label{sec:related_work}

\paragraph{3D reconstruction.}
3D reconstruction aims to recover a three-dimensional representation of a scene from a set of posed images.
Traditionally, this was accomplished with structure-from-motion~\cite{schonberger16structure-from-motion,agarwal09building,wang24vggsfm:}, multi-view stereo~\cite{schonberger16pixelwise,furukawa10accurate}, and mesh surface extraction methods~\cite{kazhdan06poisson,lorensen87marching}.
Recently, methods based on neural fields~\cite{mildenhall20nerf:,mildenhall22nerf,verbin22ref-nerf:,barron22mip-nerf,muller22instant,wang21neus:,yariv21volume} have focused on extracting both shape and appearance by learning the scene radiance function.
Through volumetric rendering and photometric losses, they are able to implicitly reconstruct a 3D density volume from which a mesh can be extracted.
However, these methods infer the parameters of each scene via test-time optimization, making them slow.
Authors have thus introduced new feed-forward reconstruction networks, generically called Large Reconstruction Models (LRMs)~\cite{wei24meshlrm:,xie24lrm-zero:,zhang24gs-lrm:,hong24lrm:}.
LRMs use transformers to predict 3D representations from a small set of input images.
Our \themethod predicts not just geometry and radiance, but also material properties and illumination.

\paragraph{Material reconstruction.}
3D reconstruction can typically recover a scene's geometry and radiance under fixed illumination.
However, it is also necessary to recover material properties so the 3D assets look realistic in novel virtual scenes.
To do so, some works focus on optimizing explicit representations such material textures on meshes~\cite{hasselgren22nvdiffrecmc,munkberg22extracting,sun23neuralpbir}, while other works have added physically-based rendering models into neural fields~\cite{boss2021nerd,boss2021neuralpil,mai23neural,Jin2023TensoIR,zarzar23splitnerf:,liang23envidr:} and Gaussian splats~\cite{liang23gs-ir:,jiang23gaussianshader:}.
While they are able to obtain high quality material reconstructions, all of these works require costly per-scene optimization.
Some have addressed this issue by learning learning feed-forward reconstructors of PBR assets~\cite{siddiqui24meta}. 
However, these are typically trained with synthetic data due to the complexity of capturing real-world material properties.
We bridge the gap between synthetic and real data by training \themethod on real data.

\paragraph{Illumination prediction.}
A key to accurate material reconstruction is accurate modelling of the scene's illumination.
Single-scene optimization methods achieve this by optimizing an internal representation of illumination using light probe images~\cite{munkberg22extracting,hasselgren22nvdiffrecmc,xiuming21nerfactor:}, spherical harmonics~\cite{liang23gs-ir:}, spherical gaussians~\cite{Jin2023TensoIR,physg2021}, or MLPs~\cite{zarzar23splitnerf:,mai23neural,liang23envidr:}.
They rely on physical models of lighting to aid the optimization.
Others have attempted to predict the illumination of a scene from a single image~\cite{diffusionLight2024,wang22stylelight}.
They rely on generative models to obtain a mapping from the input view to a possible distribution of panorama views representing the environment illumination.
Our \themethod lies in between these two lines of work:
it is similar to the predictive approach as it predicts illumination from a set of four input views, but it uses physically-based rendering losses for supervision similar to single-scene optimization methods.

\section{Methodology}%
\label{sec:Methodology}

\begin{figure*}[t]
\centering
\includegraphics[width=\linewidth]{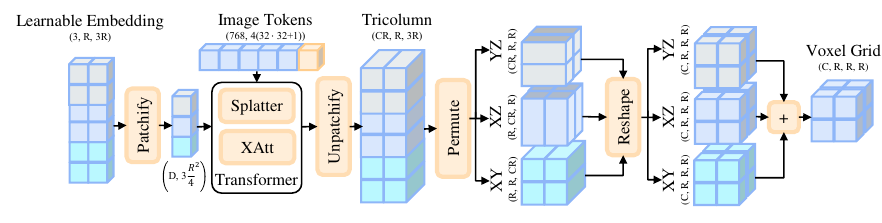}
\caption{
\textbf{Tricolumn architecture.}
We present an overview of how the novel tricolumn representation is leveraged in our \themethod.
The transformer module utilizes image tokens and a learnable embedding to predict the tricolumn representation consisting of a grid of $3R^2$ feature vectors of dimensionality $\mathbb{R}^{CR}$.
These are then split into three axis-aligned column grids, each with dimensionality $\mathbb{R}^{CR}$.
Finally, all axis-aligned column grids are reshaped into voxel grids of size $RxRxR$ with dimensionality $\mathbb{R}^{C}$ and summed into a single voxel grid.
}
\vspace{-1em}
\label{fig:tricolumn}
\end{figure*}

\newcommand{\p}{\boldsymbol{p}}
\newcommand{\bL}{\mathbf{L}}
\newcommand{\bomega}{\boldsymbol{\omega}}

We introduce \themethod, a model that reconstructs the shape, materials, and illumination of real-life 3D objects from a small set of posed input views.
\themethod equips LightplaneLRM~\cite{cao2024lightplane}, an enhanced version of LRM~\cite{hong24lrm:}, with predictors of the object's material properties and scene illumination, and with a differentiable physically-based renderer.
In \cref{sec:preliminaries}, we start by reviewing the reconstructor and the PBR model we use.
Then, in \cref{sec:tricolumn,sec:pbr_lrm,sec:illumnation}, we  describe the main building blocks of \themethod.
Finally, in \cref{sec:training_protocol}, we detail the training protocol including the losses used to supervise \themethod on real data.

\subsection{Preliminaries}%
\label{sec:preliminaries}

\paragraph{Physically-based rendering.}%
\label{sec:pbr_shader}

In order to supervise our model with a photometric loss, it is necessary to \emph{render} the reconstructed scene into an image.
The color of each pixel is equal to the \emph{radiance}
$
\bL_o 
$
emitted by the imaged object point $\p\in\mathbb{R}^3$ in the direction of the camera.
With the knowledge of a scene's illumination, geometry, and material properties, it is possible to estimate the outgoing radiance at any such point according to the \emph{reflectance equation}:
\begin{equation}
\label{eq:reflectance}
\bL_o 
= 
\int\limits_\Omega 
\left(
    k_d 
    \frac{\mathbf{a}}{\pi} 
    + 
    \mathbf{f}_s
\right)
\bL_i \,
\langle \bomega_i, \mathbf{n} \rangle \,
d\bomega_i.
\end{equation}
Here, $\bL_i$ is the incoming radiance, expressing the scene illumination falling at the imaged point $\p$.
The integral is carried over a half sphere $\Omega$ of incoming directions $\bomega_i$ determined by the surface normal $\mathbf{n}$ at $\p$.
The factor in parenthesis is a model of the the object's Bidirectional Reflectance Distribution Function (BRDF) which tells how the incoming radiance reflects off the object's surface into various directions.
The first term is the \emph{diffuse component}, and is parameterized by the object's diffuse albedo $\mathbf{a}$ and a constant $k_d$, both of which do not depend on direction;
in the integral, this term thus averages all incoming radiance over the hemisphere.
The second term $\mathbf{f}_s$ is the \emph{specular component} of the BRDF, which expresses any directional effect, such as reflections or highlights.
Note that $\bL_i$ is a function of the incoming direction $\bomega_i$ and the specular BRDF $\mathbf{f}_s$ is a function of both the incoming and the outgoing directions $\bomega_i$ and $\bomega_o$.
Radiances and BRDF depend on the light frequency too, of which we only model three RGB components.
Furthermore, all the quantities depend on the point position $\p$.
We omit these details for compactness.

Due to the complex integral in \cref{eq:reflectance}, a multitude of approximations have been proposed.
We rely on the split-sum approximation~\cite{karis2013real} for image-based lighting, which leverages the scene's illumination represented as a spherical image which envelops the scene with an infinite radius.
Following~\cite{munkberg22extracting}, we use a set of differentiable mipmaps to represent the pre-integrated lighting.
The BRDF follows the Disney~\cite{karis2013real} microfacet BRDF model, and is parameterized by the albedo $\mathbf{a}$ and the metalness $m$, and roughness $r$ at each point.
We further assume that $\bL_i$ is constant in space (hence we do not account for shadows cast by the object).

In summary, in order to apply this model to render an image of the object, we require an estimate of the object 3D shape and of its albedo, metalness, and roughness at its surfaces.
We also require a spherical image representing the directional function $\bL_i$, known as an \emph{environment map}.

\paragraph{Large Reconstruction Model (LRM).}

The aim of LRM~\cite{hong24lrm:} is to map few posed views $\mathcal{I}$ of the scene to a 3D reconstruction $V$.
LRM first extracts features $y$ from the set of input views, utilizing a pre-trained DINO~\cite{oquab24dinov2:} Vision Transformer (ViT)~\cite{dosovitskiy21an-image}.
We further follow~\cite{xu24dmv3d:} and feed into the ViT the RGB images $\mathcal{I}$, and the Pl{\"u}cker embeddings~\cite{sitzmann21light} $\mathcal{I}_{\text{Pl}}$ representing the cameras.
We also feed masks $\mathcal{M}$ of the foreground object to be reconstructed,

The extracted feature tokens enter via cross-attention the transformer architecture from Diffusion Transformer (DiT)~\cite{peebles23scalable} $\Phi$ which is tasked with reconstructing the scene.
We would like to emphasize that we borrow the DiT architecture without any diffusion process.
The tokens in this transformer encode the 3D representation of the scene, realizing a mapping $V = \Phi(V_0| y)$, where $V_0$ are the learnable initial tokens.
A crucial design choice is how to construct the tokens $V$ manipulated by this transformer, which is discussed in \cref{sec:tricolumn}.

Finally, using the memory efficient rendering algorithm proposed in Lightplane~\cite{cao2024lightplane}, a target set of views is rendered from the 3D representation.
A shaded RGB image $\hat{\mathcal{I}}$, a foreground mask image $\hat{\mathcal{M}}$, and a depth image $\hat{\mathcal{I}}_{\text{d}}$ are rendered for each view and used to supervise the LRM through a combination of the following photometric losses:
\begin{align}\label{eq:photometric_losses}
\mathcal{L}_{M} &= \text{BCE}(\mathcal{M}, \hat{\mathcal{M}}), \\
\label{eq:all_data_photometric}
\mathcal{L}_{R} &= \text{LPIPS}(\mathcal{M}\mathcal{I}, \mathcal{M}\hat{\mathcal{I}}) + \|\mathcal{M}\mathcal{I} -  \mathcal{M}\hat{\mathcal{I}}\|^2, \\
\mathcal{L}_{d} &= \|\mathcal{M}\mathcal{I}_{\text{d}} - \mathcal{M}\hat{\mathcal{I}}_{\text{d}}\|,
\end{align}
where $\mathcal{I}_{\text{d}}$ is the set of ground-truth depth maps corresponding to the input images $\mathcal{I}$ and $\mathcal{M}\mathcal{I}$ denotes a masked image.

\subsection{Tricolumn representation}%
\label{sec:tricolumn}

Crucial to the \themethod design is the representation of the 3D scene, which is encoded as a set of volumetric functions expressing opacity and material properties at every point $\p \in \mathbb{R}^3$ in space.
A simple and effective option~\cite{sun22direct} to encode a volumetric map $V: \mathbb{R}^3 \rightarrow \mathbb{R}^C$ is to use a voxel grid $V \in \mathbb{R}^{C\times D \times H \times W}$.
Here, the value $V(\p) \in \mathbb{R}^C$ interpolates the voxel values at the nearest grid points.
To predict the voxel grid $V$ with the transformer of \cref{sec:preliminaries} we could assign one token to each voxel.
However, the number $WHD$ of voxels grows cubically with the resolution of the grid, which leads to prohibitive memory demands.

The alternative adopted by models like~\cite{hong24lrm:,siddiqui24meta} is to use a \emph{triplane} representation~\cite{chan22efficient}.
In this case, the volumetric function is represented by three 2D grids 
$
V_{xy}\in \mathbb{R}^{K \times H \times W},
$
$
V_{yz}\in \mathbb{R}^{K \times W \times D},
$
and
$
V_{zx}\in \mathbb{R}^{K \times D \times H}.
$
The volumetric function at $\p=(x,y,z)$ is then expressed as
\begin{equation}\label{eq:triplane}
V(x,y,z) = f(
    V_{xy}(x, y),
    V_{yz}(y, z),
    V_{zx}(z, x)
),
\end{equation}
where $f : \mathbb{R}^{K \times K \times K} \rightarrow \mathbb{R}^C$ is a point-wise mixing function (e.g., a MLP applied to the concatenation of the three $K$-dimensional feature vectors).
Paired with a transformer, this conveniently reduces the number of tokens to $WH + HD + DW$ only.
However, triplane mixes information between different voxels, which can lead to artifacts.

We thus propose a novel \emph{tricolumn} representation instead.
This is similar to triplane as it uses a plane instead of a volume of tokens; however, each feature vector is in itself a \emph{column} of corresponding voxel cells, so that voxel-specific information is represented directly and separately.

In detail, we consider a plane of feature vectors $V_{xy}\in \mathbb{R}^{(CR)\times R\times R}$.
Because this is a plane, there are only $R^2$ tokens, which are easily handled by the transformer.
However, $V_{xy}$ can be reshaped into a  $C \times R \times R \times R$ voxel grid by unflattening the first dimension.
We use the symbol $V_{xy}(x,y;z) \in \mathbb{R}^C$ to denote the feature vector obtained by interpolating this grid in 3D.

When three such planes are used, the full feature vector at $\p$ is decoded as:
\begin{equation}
\label{eq:tricolumn}
V(x,y,z)
=
f(
    V_{xy}(x,y;z),
    V_{yz}(y,z;x),
    V_{zx}(z,x;y)
).
\end{equation}
Note that this construction is strictly different from the triplane equation~\eqref{eq:triplane} as each term depends on all three spatial coordinates.
We opt for using summation followed by an MLP as the mixing function $f$.

\begin{figure}[t]
\centering
\includegraphics[width=\linewidth]{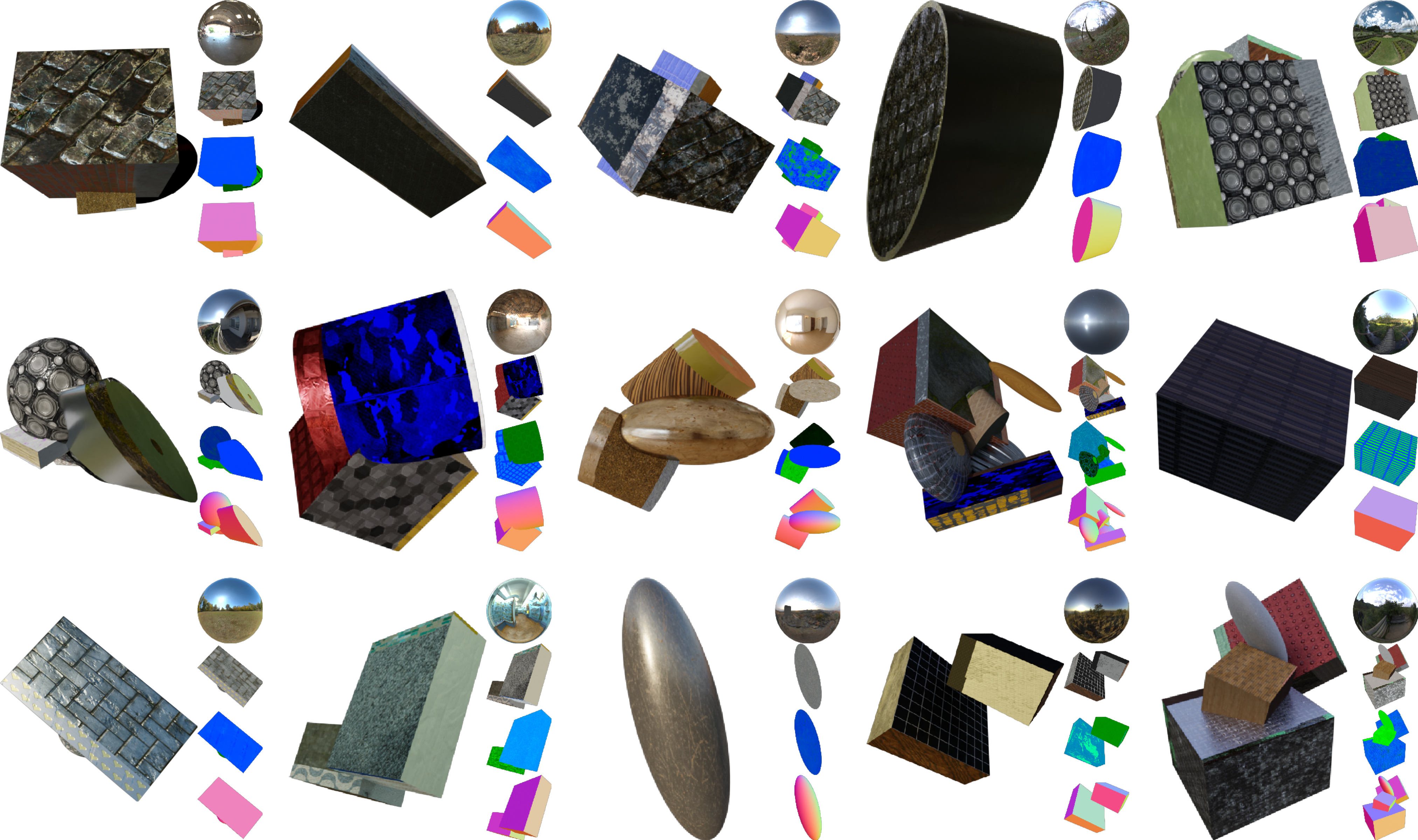}
\caption{
\textbf{Examples from our procedural dataset}
visualizing the shaded image, environment map, materials, and normals from our environment-map-endowed version of Zeroverse \cite{xie2024lrmzero}.
}%
\vspace{-1em}
\label{fig:zeroverse}
\end{figure}

\subsection{PBR reconstruction model}%
\label{sec:pbr_lrm}

The LRM models of~\cite{hong24lrm:,xu24dmv3d:} only recover the shape and radiance of the scene.
Like in~\cite{siddiqui24meta}, we extend \themethod to also recover the material properties.
We do so by extending the tricolumn representation to encode the albedo $\mathbf{a}$ (instead of radiance) and two additional channels representing the scene's metalness $m$ and roughness $r$.
We also note the importance of accurate normals for relighting.
While a normal image could be directly estimated by projecting the rendered depth image into 3D space and taking the gradient of the opacity field,
$\bar{\mathcal{I}}_{\text{n}} = \nabla \sigma(\pi^{-1}(\hat{\mathcal{I}}_{\text{d}}))$,
this estimate tends to be noisy.
Therefore, we follow previous works~\cite{verbin22ref-nerf:} and predict three additional channels representing the scene's normals.
The predicted normal image $\hat{\mathcal{I}}_{\text{n}}$ is then supervised with the pseudo-gt normals from the opacity field $\bar{\mathcal{I}}_{\text{n}}$ via:
\begin{equation}\label{eq:all_data_normal_opacity}
\mathcal{L}_{N} 
= 
1 - 
\langle 
    \mathcal{M} \bar{\mathcal{I}}_{\text{n}},  
    \mathcal{M} \hat{\mathcal{I}}_{\text{n}} 
\rangle.
\end{equation}

\paragraph{Direct supervision using synthetic data.}

Reconstruction models that predict radiance can be supervised by \emph{rendering} the predicted 3D scene into images and comparing these to training views, using a photometric loss.
In our case, however, the model does not predict radiance, but albedo, roughness, and metalness, which cannot be rendered into a `shaded' image unless illumination is also known.

A simple approach for supervising the model that does not require knowledge of illumination is to render albedo, roughness, and metalness images $\hat{\mathcal{I}}_{\text{a}}$, $\hat{\mathcal{I}}_{\text{r}}$, and $\hat{\mathcal{I}}_{\text{m}}$, and compare these to their corresponding ground-truth renders.
We use this approach to supervise our model with synthetic data by applying the following loss on the set of material renders 
$\hat{\mathcal{I}}_{\text{a}},
\hat{\mathcal{I}}_{\text{m}},
\hat{\mathcal{I}}_{\text{r}}$,
and also on the normal render $\hat{\mathcal{I}}_{\text{n}}$:
\begin{equation}\label{eq:synthetic_image}
\mathcal{L}_S = 
\ell_\mathcal{M}(\mathcal{I}_{\text{a}}, \hat{\mathcal{I}}_{\text{a}})
+ \ell_\mathcal{M}(\mathcal{I}_{\text{r}}, \hat{\mathcal{I}}_{\text{r}})
+ \ell_\mathcal{M}(\mathcal{I}_{\text{m}}, \hat{\mathcal{I}}_{\text{m}})
+ \ell_\mathcal{M}(\mathcal{I}_{\text{n}}, \hat{\mathcal{I}}_{\text{n}})
\end{equation}
where $\ell_\mathcal{M}(u, v) = \| \mathcal{M} (u - v) \|_\epsilon$
is the Huber loss $\| \cdot \|_\epsilon$ with cut-off threshold $\epsilon$ masked by $\mathcal{M}$.

\subsection{Training on real data by predicting illumination}%
\label{sec:illumnation}

The learning formulation of \cref{sec:pbr_lrm} requires ground-truth material properties to be know, which is not the case for real-world data where only shaded images are available.
In order to fine-tune our model using real data with a photometric loss, then, it is necessary to also estimate the scene illumination. 
With this, we can use the differentiable physically-based shader detailed in \cref{sec:pbr_shader} to render the shaded scene appearance, and compare that to the input images through a photometric loss.

In order to estimate the scene's illumination, we propose using an additional DiT module.
The latter takes as input the feature tokens $y$ computed by the image feature extractor and predicts a cubemap representation of the scene's lighting.
Recall that a cubemap is a function $C \rightarrow \mathbb{R}^3$ that assigns every point $\p$ of the cube surface $C = \{\p : \| \p \|_\infty = 1\}$ to the radiance $\bL_i(\bomega_i)$ incoming from direction $\bomega_i \propto -\p$, and is thus a representation of the environment illumination (this works because we assume that the function $\bL_i$ only depends on direction, but not on position).
The cubemap is unfolded to a collection of six 2D images, which are directly predicted by the DiT module.

\begin{table*}[ht]
\newcommand{\no}{\color{Maroon}{\xmark}}%
\newcommand{\yes}{\checkmark}%
\definecolor{rowcol}{RGB}{230,245,255}%
\resizebox{\textwidth}{!}{
\setlength{\arraycolsep}{1pt}
\begin{tabular}{clcccccccccccc}
\toprule
 & & \multicolumn{4}{c}{Novel Scene Relighting} & \multicolumn{4}{c}{Novel View Synthesis} & \multicolumn{3}{c}{Geometry Estimation} & Runtime \\ \midrule
& Method & \multicolumn{1}{l}{PSNR-H↑} & \multicolumn{1}{l}{PSNR-L↑} & \multicolumn{1}{l}{SSIM↑} & \multicolumn{1}{l}{LPIPS↓} & \multicolumn{1}{l}{PSNR-H↑} & \multicolumn{1}{l}{PSNR-L↑} & \multicolumn{1}{l}{SSIM↑} & \multicolumn{1}{l}{LPIPS↓} & \multicolumn{1}{l}{Depth↓} & \multicolumn{1}{l}{Geometry↓} & \multicolumn{1}{l}{Shape↓} & \multicolumn{1}{l}{Time (hr.)↓} \\ \midrule
\multirow{7}{*}{\rotatebox[origin=c]{90}{Single-Scene Opt.}}
& PhySG~\cite{physg2021}            & 21.81 & 28.11 & 0.960 & 0.055 & 24.24 & 32.15 & 0.974 & 0.047 & 1.90 & 0.17 & 9.28 & $\sim$3 \\
& NVDiffRec~\cite{munkberg22extracting}        & 22.91 & 29.72 & 0.963 & 0.039 & 21.94 & 28.44 & 0.969 & 0.030 & \underline{0.31} & \underline{0.06} & 0.62 & $\sim$1 \\
& NeRD~\cite{boss2021nerd}             & 23.29 & 29.65 & 0.957 & 0.059 & 25.83 & 32.61 & 0.963 & 0.054 & 1.39 & 0.28 & 13.70 & \textgreater{}20 \\
& NeRFactor~\cite{xiuming21nerfactor:}        & 23.54 & 30.38 & 0.969 & 0.048 & 26.06 & 33.47 & 0.973 & 0.046 & 0.87 & 0.29 & 9.53 & \textgreater{}20 \\
& InvRender~\cite{zhu2022learning}        & 23.76 & 30.83 & 0.970 & 0.046 & 25.91 & 34.01 & 0.977 & 0.042 & 0.59 & 0.06 & \underline{0.44} & $\sim$3 \\
& NVDiffRecMC~\cite{hasselgren22nvdiffrecmc}      & \underline{24.43} & \underline{31.60} & \underline{0.972} & \underline{0.036} & \underline{28.03} & \underline{36.40} & \underline{0.982} & \underline{0.028} & 0.32 & \textbf{0.04} & 0.51 & $\sim$2 \\
& Neural-PBIR~\cite{sun23neuralpbir}      & \textbf{26.01} & \textbf{33.26} & \textbf{0.979} & \textbf{0.023} & \textbf{28.82} & \textbf{36.80} & \textbf{0.986} & \textbf{0.019} & \textbf{0.30} & 0.06 & \textbf{0.43} & {$\sim$1} \\
\midrule
\multirow{4}{*}{\rotatebox[origin=c]{90}{Predictive}}
& MetaLRM~\cite{siddiqui24meta} & 21.50 & 28.06 & 0.957 & 0.052 & 19.55 & 25.98 & 0.956 & 0.051 & 8.669 & \underline{0.06} & 1.26 & {0.004} \\
& w/o ZeroVerse & 23.53 & 30.31 & 0.965 & 0.048 & \underline{21.53} & 28.78 & 0.965 & 0.048 & 1.126 & 0.09 & 0.74  & 0.006 \\
& w/o Shading Loss & \underline{23.89} & \underline{30.85} & \underline{0.967} & \underline{0.042} & 21.42 & \underline{29.47} & \underline{0.967} & \underline{0.041} & \underline{0.762} & \textbf{0.06} & \textbf{0.30} & 0.004 \\
& \textbf{\themethod (ours)} & \textbf{24.55} & \textbf{31.61} & \textbf{0.969} & \textbf{0.038} & \textbf{21.70} & \textbf{29.62} & \textbf{0.968} & \textbf{0.037} & \textbf{0.709} & 0.06 & \underline{0.37} & 0.006  \\ \bottomrule
\end{tabular}
}
\caption{
\textbf{Relighting evaluation on StanfordORB~\cite{kuang2023stanfordorb}} comparing our method to other feedforward predictors of PBR-textured assets (two ablations and MetalLRM).
For completeness, we also include single-scene optimization methods in the top part of the table.
Our model outperforms most single-scene optimization baseline methods in terms of novel scene relighting and shape estimation despite being a predictive model and having access to only four input views.
More impressively, it only requires a few seconds to reconstruct an object's mesh with PBR materials, while single-scene optimization methods have runtimes in the order of hours.
}%
\vspace{-1em}
\label{tab:main_results}
\end{table*}%
\let\yes\undefined%
\let\no\undefined%

\paragraph{Supervising illumination with synthetic data.}

In synthetic data, the cubemap is known and its predictor can be supervised directly with the environment-map loss
\begin{equation}\label{eq:env_loss}
\mathcal{L}_\text{env} = 
\| \mathcal{I}_\text{e} - \hat{\mathcal{I}}_{\text{e}} \|^2
+ LPIPS(\mathcal{I}_{\text{e}}, \hat{\mathcal{I}}_{\text{e}}),
\end{equation}
comprising the photometric loss from~\eqref{eq:synthetic_image} using the predicted cubemap projected into an equirectangular image $\hat{\mathcal{I}}_{\text{e}}$ and the ground-truth equirectangular image $\mathcal{I}_\text{e}$.
In practice, we use a procedurally generated dataset of synthetic assets detailed in \cref{sec:training_protocol} to provide ground-truth illumination.

\paragraph{Supervising illumination with real data.}

In real data, the cubemap is observed indirectly via the effect it has on the object appearance according to the reflectance equation~\eqref{eq:reflectance}, which provides only weak supervision.
We note, however, that the \emph{scene background} visible in the input images is, to a first approximation, equal to the function $\bL_i$, and can thus provide direct supervision of the latter.
We can \emph{render} the background observed from any given camera by assigning to each pixel the color
$
\bL_i(\bomega_i) = C(\bomega_i / \|\bomega_i\|_\infty)
$
of the incoming radiance from the direction $\bomega_i$ of the pixel's ray.
Since input images are in sRGB color space, the predicted background image is first tone-mapped and then composited with the foreground object prediction to obtain a scene image.
The latter allows us to supervise both the illumination prediction and the shaded object prediction through the photometric loss $\mathcal{L}_{R}$ of \cref{eq:all_data_photometric}.
Here, as training data, we use abundant object-centric video captures from CO3Dv2~\cite{reizenstein21common} as described in the next section.

\subsection{Training protocol}\label{sec:training_protocol}

\themethod's training minimizes the total loss
\begin{equation}\label{eq:total_loss}
\mathcal{L}_\text{tot} = \mathcal{L}_N + \mathcal{L}_{S} + \mathcal{L}_\text{TV} + \mathcal{L}_d + \mathcal{L}_M + \mathcal{L}_R + \mathcal{L}_{\text{TV}},
\end{equation}
where
$
\mathcal{L}_{\text{TV}}
=
\text{TV}(\hat{\mathcal{I}}_{\text{n}}) +
\text{TV}(\hat{\mathcal{I}}_{\text{a}}) +
\text{TV}(\hat{\mathcal{I}}_{\text{d}})
$
is the Total-Variation (TV) regularization on normal, albedo, and depth images promoting smooth solutions.
$\mathcal{L}_\text{tot}$ is optimized with the Adam~\cite{kingma14adam:} with learning rate $10^{-4}$ until convergence.
As mentioned before, individual terms of $\mathcal{L}_\text{tot}$ are switched on or off depending on whether a real or a synthetic training example is present in the batch of training images.

\paragraph{Training data.}

Following~\cite{siddiqui24meta}, \themethod is trained on a large dataset of 140k artist-created assets similar to Objaverse.
However, this dataset often lacks high-quality material maps where, for instance, distinguishing between missing material images and material images with zero values is impossible.
Moreover, there is a big domain gap between the synthetic objects and the real ones we intend to twin.

As such, together with the artist-created assets, we also train on two more data sources:
1) A large dataset of procedurally generated assets with clean PBR textures and environment maps; 
2) real-object captures of CO3Dv2~\cite{reizenstein21co3d}. 
Both datasets are detailed next.

\paragraph{Endowing Zeroverse with environment maps.}

Following LRM-Zero~\cite{xie2024lrmzero}, we generate a set of 10k randomized 3D objects with realistic material properties sampled from the MatSynth~\cite{vecchio2024matsynth} dataset.
Additionally, to enable the environment-map supervision loss $\mathcal{L}_\text{env}$~\eqref{eq:env_loss}, for each rendered scene, we randomly sample one of $\sim$1k environment maps from PolyHaven~\cite{PolyHaven}.
We render color, normal, material, depth, and foreground mask images for each object from a predetermined set of viewpoints (see examples in \cref{fig:zeroverse}).
As such, samples from this dataset admit optimizing all terms of the total loss $\mathcal{L}_\text{tot}$, including the environment loss $\mathcal{L}_\text{env}$, which is crucial for stable training of \themethod.

\paragraph{Exploiting real object captures.}

We also train on CO3Dv2~\cite{reizenstein21co3d}, which is a large-scale real-world dataset of ``turn-table'' videos capturing objects of common categories.
Each frame in the dataset comes with camera and foreground mask annotations.
As such, for each CO3Dv2 sample, we can optimize the normal, mask, photometric, and TV losses $\mathcal{L}_\text{tot}^\text{real} = \mathcal{L}_N + \mathcal{L}_M + \mathcal{L}_R + \mathcal{L}_\text{TV}$.
Since $\mathcal{L}_\text{tot}^\text{real}$ is a function of the illumination image $\hat{\mathcal{I}}_\text{e}$ and PBR renders $\hat{\mathcal{I}}_\text{a}$, $\hat{\mathcal{I}}_\text{m}$, $\hat{\mathcal{I}}_\text{r}$, the loss provides a valuable supervisory signal for all volumetric predictors.
Furthermore, using real training data narrows the domain gap between the training set and the target domain of \themethod, which are real images.
\begin{figure}[ht]
\centering
\includegraphics[width=\linewidth]{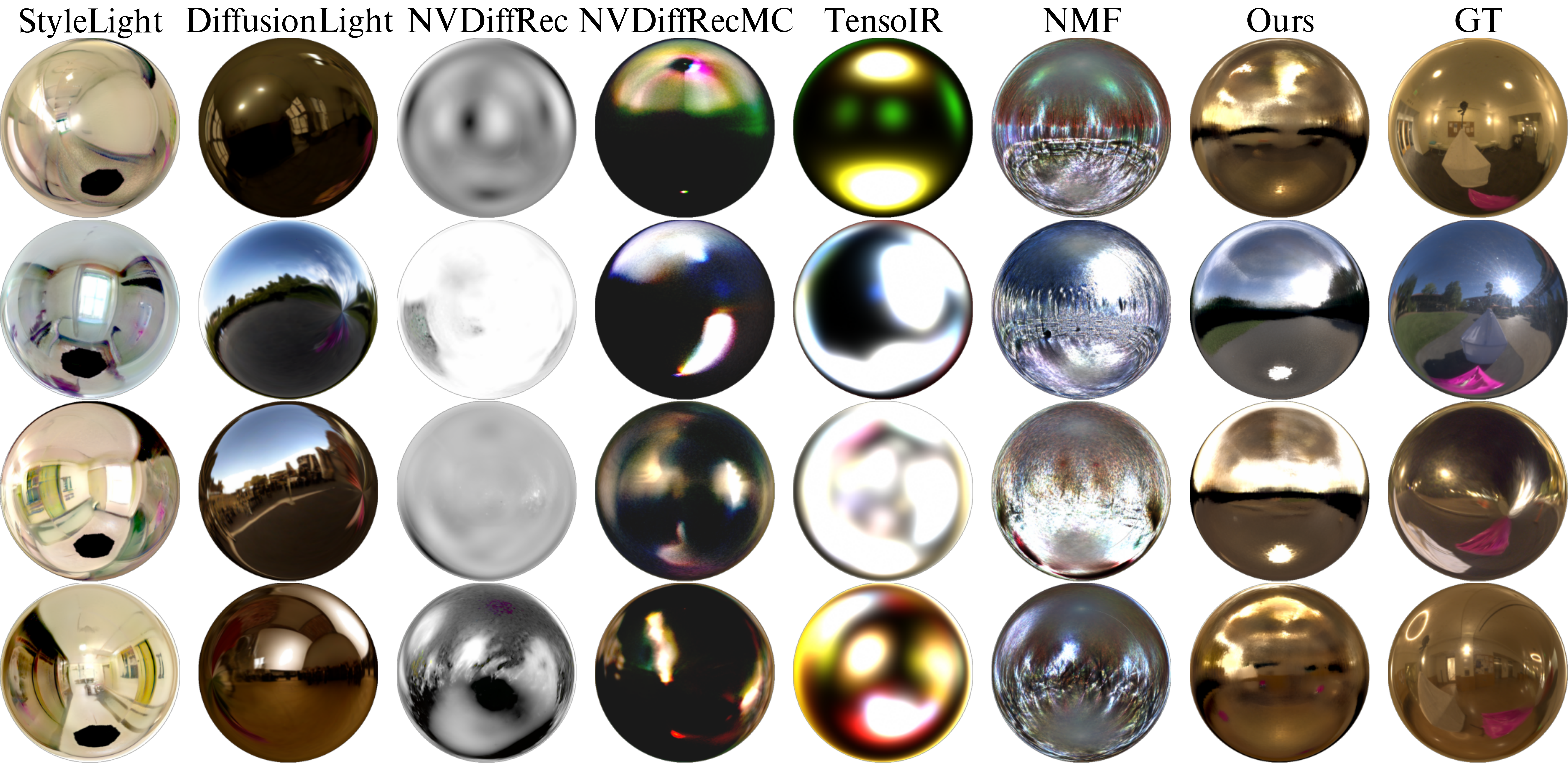}
\caption{
\textbf{StanfordORB Illumination Visualization.} We visualize the illumination estimation of multiple baselines and our method on four different scenes from the StanfordORB dataset.
}%
\vspace{-1em}
\label{fig:qualitative_illumination}
\end{figure}

\begin{table}[ht]
\resizebox{\linewidth}{!}{
\begin{tabular}{@{}lccccc@{}}
\toprule
Method & \multicolumn{1}{l}{Angular Error  ↓} & \multicolumn{1}{l}{RMSE↓} & \multicolumn{1}{l}{Norm. RMSE↓} & \multicolumn{1}{l}{SI RMSE↓} & \multicolumn{1}{l}{Time (hr.)↓} \\ \midrule

NMF~\cite{mai23neural}& \textbf{11.24} & \textbf{0.487} & \textbf{0.524} & \underline{0.394} & $\sim$ 2 \\
TensoIR~\cite{Jin2023TensoIR} & 15.08 & 0.752 & 0.798 & 0.451 & $\sim$ 2 \\
NVDiffRec~\cite{munkberg22extracting}& \underline{12.26} & 0.723 & 0.715 & \textbf{0.367} & $\sim$ 1 \\
NVDiffRecMC~\cite{hasselgren22nvdiffrecmc} & 13.25 & \underline{0.574} & \underline{0.614} & 0.527 & $\sim$ 2 \\
\midrule
StyleLight~\cite{wang22stylelight} & 13.05 & 0.640 & 0.689 & \textbf{0.440} &  0.011 \\
DiffusionLight~\cite{diffusionLight2024} & \underline{11.35} & \underline{0.581} & \textbf{0.569} & 0.544 & 0.317 \\
\textbf{\themethod (ours)} & \textbf{8.09} & \textbf{0.570} & \underline{0.602} & \underline{0.472} & \textbf{0.006} \\ \bottomrule
\end{tabular}
}
\caption{
     \textbf{Illumination prediction results.} 
     We provide a quantitative evaluation of the illumination estimation for  baselines and our \themethod.
     Rows are grouped by single-scene optimization models (top) and predictive feed-forward models (bottom).
}%
\vspace{-1em}
\label{tab:illumination_results}
\end{table}


\begin{figure}[t]
\centering
\includegraphics[width=\linewidth]{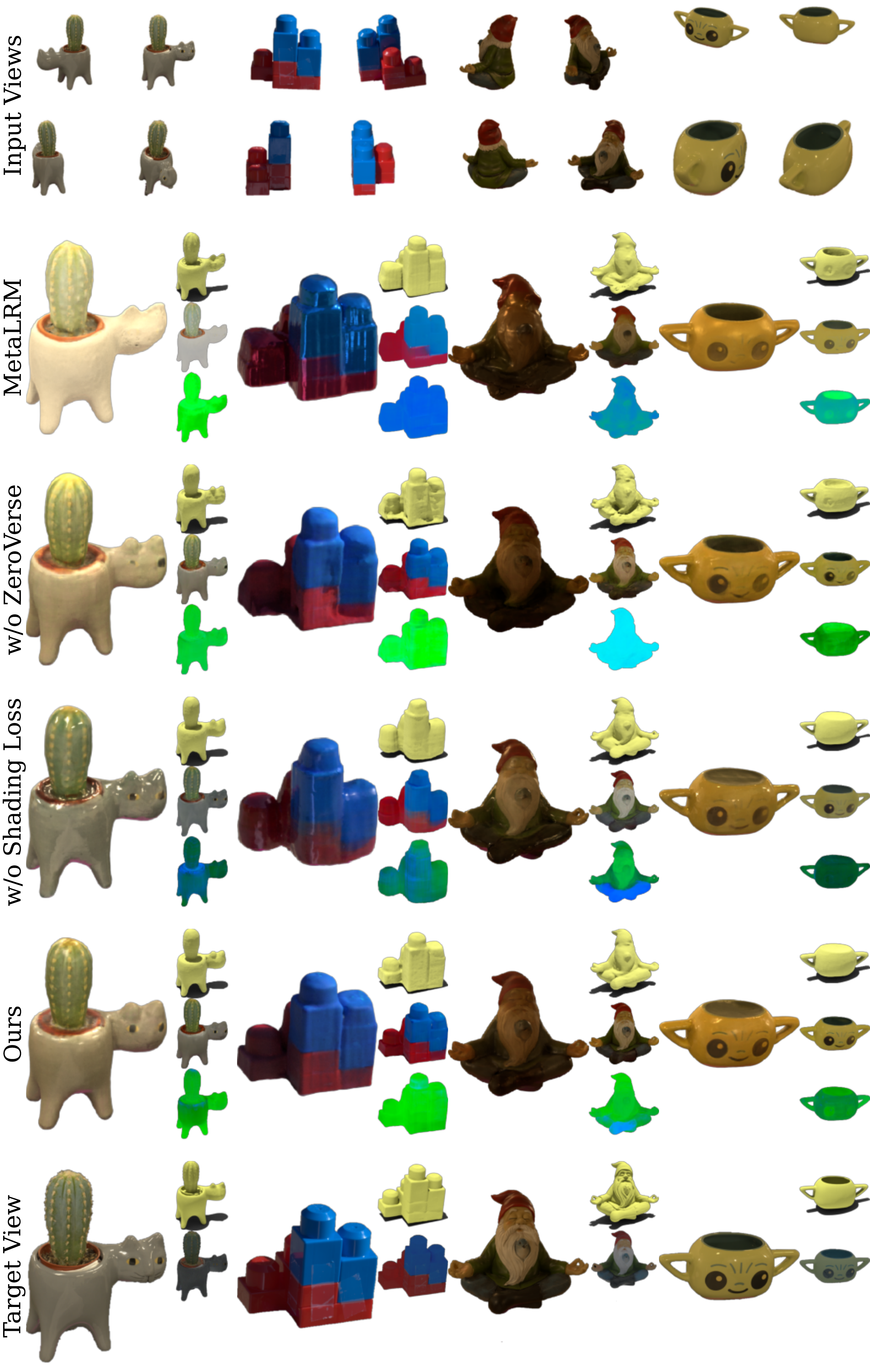}
\caption{
\textbf{Qualitative results on StanfordORB\@.}
We visualize the shaded RGB, geometry reconstruction, and material predictions for our baseline, two ablations, and our \themethod. 
Our model is able to better predict relightable objects thanks to the combination of curated synthetic training data and the shading loss on real data.
}%
\vspace{-1em}
\label{fig:qualitative_stanfordorb}
\end{figure}

\section{Experiments}%
\label{sec:experiments}

\begin{figure*}[htb]
\centering
\includegraphics[width=\textwidth]{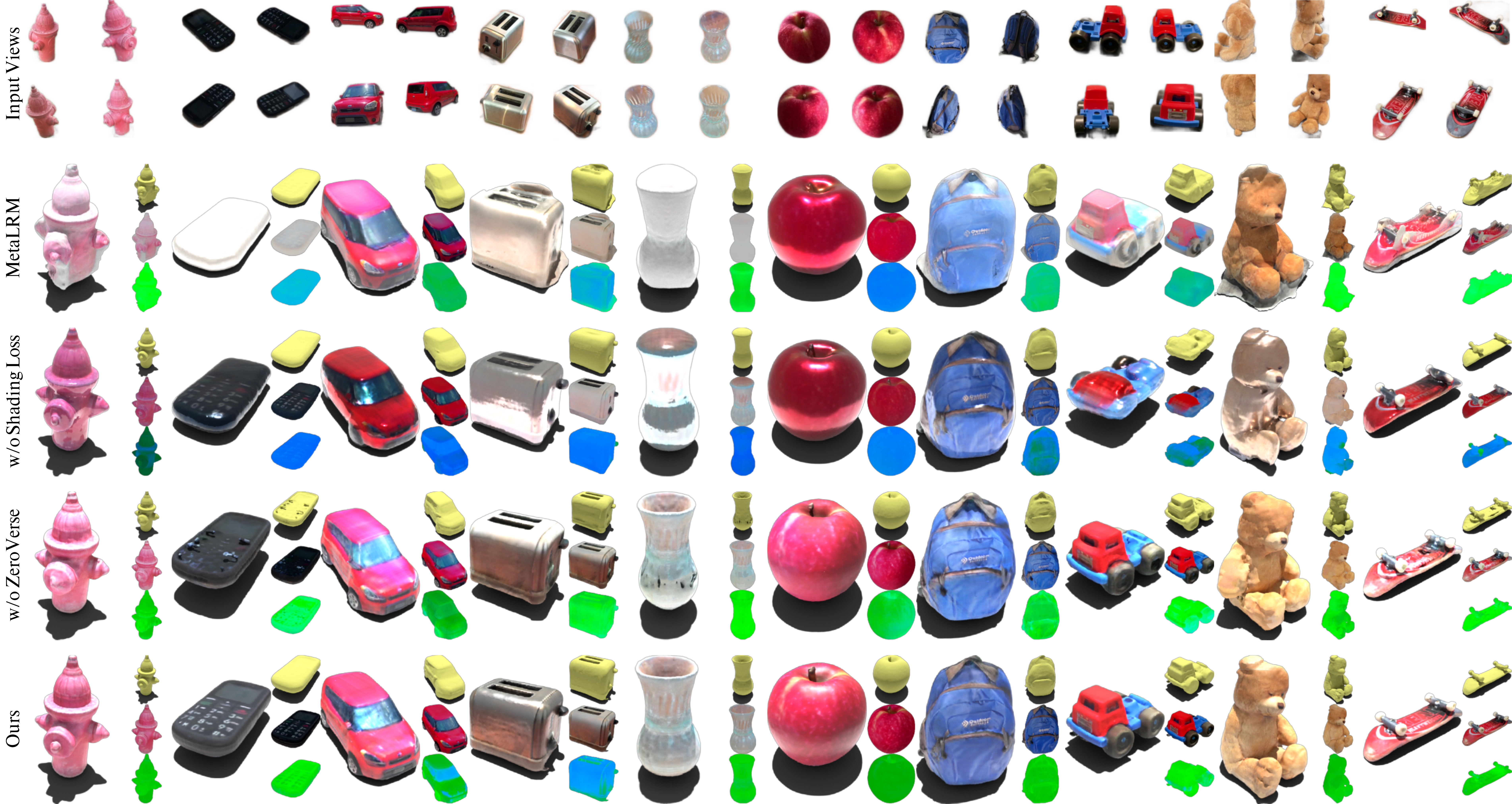}
\caption{
\textbf{CO3Dv2 Qualitative Results.}
We visualize predicted PBR meshes from the CO3Dv2 dataset illuminated under direct sunlight as well as their geometry, albedo, and material properties.
The shading loss aids at predicting more realistic material properties, while the synthetic data improves the predicted geometry and albedo.
}%
\vspace{-1em}
\label{fig:qualitative_co3d}
\end{figure*}

\paragraph{Evaluation data.}

We evaluate \themethod on the StanfordORB~\cite{kuang2023stanfordorb} dataset which comprises 42 captures of 14 different objects each imaged under 3 different illuminations.
Each object has a high quality geometric reconstruction obtained using a 3D scanner, and comes with ground-truth camera poses and estimated illumination.

\paragraph{Evaluation protocol.}

For each evaluation scene, we select four views from the set of training images as input frames via furthest camera-origin point sampling.
With such input views, \themethod reconstructs a 3D PBR-textured mesh.
Each predicted mesh is rendered with Blender~\cite{blender2018} to held-out test views while being lit with a given ground-truth environment map.
Following~\cite{kuang2023stanfordorb}, the renders are compared to the corresponding ground-truth images using scale-invariant versions of LPIPS~\cite{zhang18the-unreasonable}, SSIM~\cite{wang04bimage}, and PSNR\@.

\paragraph{Baselines.}

Our main comparison focuses on existing feedforward predictors of PBR-textured meshes.
As such, we selected the state-of-the-art MetalLRM few-view reconstructor from AssetGen~\cite{siddiqui24meta} as our main baseline.
We use the version of MetalLRM which is conditioned on four shaded input views as opposed to the albedo+shaded conditioning used in the AssetGen pipeline.
We also compare to two ablations of our method.
The first ablation, \textbf{\themethod w/o Shading Loss}, evaluates the contribution of training on real data and omits the training images of CO3Dv2.
Second, we compare to \textbf{\themethod w/o Zeroverse} which ablates training of the environment-map and of the PBR-texture predictor on our version of the procedurally-created Zeroverse.
For completeness we also include several single-scene optimization methods which ``overfit'' a relightable implicit shape in each scene from scratch.
Note that, in order to save GPU memory, our feedforward predictors randomly sample only four source views from the pool of all available training images of each scene.
This puts us at a disadvantage w.r.t.~single-scene optimizers as they leverage all available scene views for each reconstruction.

\paragraph{Relighting evaluation.}

All quantitative results are reported in \cref{tab:main_results}.
\themethod is the best among feedforward reconstructors outperforming its own two ablations and MetalLRM\@.
For instance, \themethod achieves a 27\% reduction in both novel-scene relighting and novel-view synthesis LPIPS w.r.t. the strongest baseline (MetaLRM), while maintaining comparable inference time.
Remarkably, our method is on par with most single-scene optimization methods while being significantly faster with few seconds per prediction compared to several hours of the optimization-based methods.
Examples of our predictions compared to baselines are in \cref{fig:qualitative_stanfordorb}.
Additionally, we evaluate qualitatively on testing sequences of CO3Dv2 in \cref{fig:qualitative_co3d}.

\paragraph{Illumination evaluation.}

We evaluate \themethod's ability to predict a scene's illumination from four input images.
Since there are no existing baselines for few-view conditioned feed-forward illumination prediction, we compare to environment map estimates output by the single-scene optimizers in \cref{tab:main_results}, and to single-view conditioned predictors.
We follow~\cite{diffusionLight2024} and evaluate the illumination quality in terms of angular error, RMSE, normalized RMSE, and scale-invariant RMSE\@.
Results are in~\cref{tab:illumination_results,fig:qualitative_illumination}.
\begin{table}[t]
\centering\small%
\begin{tabular}{@{}lccc@{}}
\toprule
3D representation  & \multicolumn{1}{l}{PSNR↑} & \multicolumn{1}{l}{SSIM↑} & \multicolumn{1}{l}{LPIPS↓} \\ \midrule
Triplane  & 31.20  & 0.972	 & 0.037  \\ 
\textbf{Tricolumn (ours)}  & \textbf{32.81}  & \textbf{0.966} & \textbf{0.029}  \\ \bottomrule 
\end{tabular}
\caption{
\textbf{Tricolumn ablation}
evaluating novel view synthesis on the StanfordORB dataset for a LightplaneLRM \cite{cao2024lightplane} using the triplane 3D representation or our proposed tricolumn instead.
Our tricolumn outperforms triplane on all reconstruction metrics.
}%
\vspace{-1em}
\label{tab:tricolumn_results}
\end{table}
\paragraph{Tricolumn ablation.}
We also evaluate the effect of the proposed tricolumn representation.
In \cref{tab:tricolumn_results}, we report the novel-view synthesis performance of the vanilla LightplaneLRM~\cite{cao2024lightplane} model trained only on our synthetic dataset, and utilizing either tricolumn or triplane representations.
Tricolumn outperforms triplane significantly by 22\% in the most informative LPIPS metric, validating its contribution.
\section{Conclusion}%
\label{sec:Conclusion}

We have presented \themethod, a new LRM that can recover the creation of high-quality 3D assets with accurate geometry, material properties, and illumination from only a few posed images in a feed-forward manner.
\themethod leverages the novel tricolumn representation, whose memory-efficiency admits processing of voxel grids with large transformers.
For training, we showed the power of generating synthetic PBR assets procedurally, addressing the scarcity of high-quality PBR assets in existing synthetic datasets.
We also showed that, by predicting illumination, the model can learn from real-life datasets without ground truth illumination and material properties.
We evaluated \themethod on the StanfordORB benchmark, outperforming existing feed-forward predictors and achieving comparable quality to optimization-based methods while being orders of magnitude faster.
\appendix
\clearpage
\setcounter{page}{1}
\maketitlesupplementary

\section{Supplementary Material}
\label{sec:suppl}

\subsection{Network Details}%
\paragraph{Tricolumn Encoder}%
We predict a tricolumn grid with spatial dimensions of $R=128$, and a feature dimension of $D=1024$ from our transformer network.
The transformer network consists of 16 cross attention layers, interlaced with splatting layers before cross attention layers number $0$, $8$, and $15$.
\paragraph{Illumination Module}%
The illumination module consists of a transformer with 16 cross-attention layers using a feature dimension of $D=1024$.
The predicted cubemap has a resolution of $128\times128$ for each face.
We use as positional encoding the 3D directional vectors corresponding to each pixel in the cubemap.
These directional vectors are then passed to a harmonic encoding in order to produce a $D$-dimensional positional encoding.

\subsection{Training Details}%
We train using input images of resolution $256 \times 256$, while the evaluation is performed at a resolution of $512 \times 512$ following Stanford ORB~\cite{kuang2023stanfordorb}.
As mentioned in the main script, we use a combination of multiple loss functions to supervise \themethod during training.
Specifically, we use the losses $\mathcal{L}_{S}$ to supervise material properties and illumination with synthetic data, $\mathcal{L}_N$ to supervise predicted normals using our pseudo-gt normals from density, $\mathcal{L}_\text{TV}$ to smoothen albedo, normals, and depth, $\mathcal{L}_d$ to supervise the rendered depth, $\mathcal{L}_M$ to supervise the foreground object's opacity mask, and $\mathcal{L}_R$ to supervise the whole model using the composite rendered images.
The final loss we use is calculated as follows:
\begin{equation}\label{eq:total_loss_coeffs}
\mathcal{L}_\text{tot} = \mathcal{L}_R + \mathcal{L}_{S} +  \mathcal{L}_M + 0.1 \mathcal{L}_N  + 0.1 \mathcal{L}_\text{TV} + 0.1 \mathcal{L}_d.
\end{equation}
We finetune our model until convergence using 32 A100 GPUs, which takes approximately two days.

\subsection{BRDF details}
As mention in the main script, we are able to compute the outgoing radiance at any point according to the reflectance equation \cref{eq:reflectance}.
We utilize the disney BRDF parameterized by albedo, metalicity and roughness together with the split-sum approximation as per~\cite{karis2013real}.
That is, we compute the diffuse and specular components of the reflectance equation $\mathbf{\hat{L}}_d$ and $\mathbf{\hat{L}}_s$ as
\begin{equation}
    \mathbf{\hat{L}}_d =   \frac{\mathbf{\hat{k}}_d \mathbf{\hat{a}}}{\pi} \hat{L}^i_d , \quad \mathbf{\hat{L}}_s =  (\mathbf{\hat{F}}_r * F_1 + F_2) \hat{L}^i_s   ,
\label{eq:specular_approx2}
\end{equation}
where $F_1$ and $F_2$ are precomputed and stored in a 2D look-up table, $\hat{L}^i_s$ and $\hat{L}^i_d$ are pre-integrated illumination maps which we obtain from our predicted illumination cubemap in a differentiable manner following~\cite{munkberg22extracting}, and where

\vspace{-.5cm}
\begin{equation}
\begin{split}
    & \mathbf{\hat{k}}_d = (1 - \hat{m}) * (1 - \mathbf{\hat{F}}_r). \\
     & \mathbf{\hat{F}}_r = \mathbf{\hat{F}}_0 + (1 - \hat{\rho} - \mathbf{\hat{F}}_0) * (1 - \langle \mathbf{\hat{n}}, \mathbf{\omega}_o \rangle )^5, \\
     & \mathbf{\hat{F}}_0 = (1 - \hat{m}) * 0.04 + \hat{m} * \mathbf{\hat{a}}.
\label{eq:diffuse_fresnel}
\end{split}
\end{equation}

\subsection{Meshing Details}
We first estimate a point cloud covering the object by rendering a set of depth images from fixed viewpoints and projecting them into 3D space.
This point cloud is used to initialize a tetrahedral grid with a resolution of $128^3$.
We then sample density values from our implicit representation composed of a predicted tricolumn and the learnt lightplane MLP at the tetrahedral vertex locations.
In order to remove some of the noise from the density estimates we apply a graph-based smoothing operator along the edges of the tetrahedral grid.
Finally, we apply marching tetrahedra on the density grid to extract a mesh.

Once a mesh has been extracted we are then required to extract PBR textures.
We first simplify the mesh through edge contraction with a target of $40$K faces.
Since we are targeting single object reconstruction we then filter the mesh to keep only the single largest cluster of faces thus removing any floating artifacts which could have been created due to noisy density.
Given this simplified mesh, we then perform UV mapping and extract albedo, metalicity, and roughness for each uv index using its corresponding 3D point on the mesh.
Given the 3D surface point and corresponding normal, we move a fixed distance along the normal direction and perform volume rendering to extract the albedo, metalicity, and roughness in order to account for distributed density along the surface of the object.

After extracting a PBR textured mesh, we then apply the texture enhancement proposed int~\cite{siddiqui24meta} in order to refine the predicted PBR textures.

\subsection{Visualization Videos}
We provide additional visualization videos of reconstructions generated with \themethod together with this supplementary material.
We visualize the predictions from our model under direct sunlight for four different instances of a wide range of object categories in CO3Dv2 ranging from simple shapes like balls to complex objects like motorcycles.
Our model is able to provide good quality relightable reconstructions thanks to the combination of real and synthetic data used during training.

{
    \small
    \bibliographystyle{ieeenat_fullname}
    \bibliography{main,vedaldi_general,vedaldi_specific}
}
\end{document}